\documentclass[sigconf]{acmart}

\usepackage{multirow}

\usepackage{algorithm}
\usepackage{algorithmic}
\usepackage{multirow}
\usepackage{newfloat}
\usepackage{listings}

\usepackage{multirow} 

\usepackage{makecell}

\AtBeginDocument{%
  \providecommand\BibTeX{{%
    \normalfont B\kern-0.5em{\scshape i\kern-0.25em b}\kern-0.8em\TeX}}}
\setcopyright{acmcopyright}
\copyrightyear{2023}
\acmYear{2023}
\setcopyright{acmlicensed}\acmConference[WWW '23]{Proceedings of the ACM
	Web Conference 2023}{May 1--5, 2023}{Austin, TX, USA}
\acmBooktitle{Proceedings of the ACM Web Conference 2023 (WWW '23), May
	1--5, 2023, Austin, TX, USA}
\acmPrice{15.00}
\acmDOI{10.1145/3543507.3583205}
\acmISBN{978-1-4503-9416-1/23/04}

\begin{document}
\settopmatter{printacmref=false} 

\title{FormerTime: Hierarchical Multi-Scale Representations for Multivariate Time Series Classification}

\author{Mingyue Cheng$^{1}$, Qi Liu$^{1\ast}$, Zhiding Liu$^{1}$,  Zhi Li$^{2}$, Yucong Luo$^{1}$, Enhong Chen$^{1}$}

\affiliation{%
	\institution{$^1$Anhui Province Key Laboratory of Big Data Analysis and Application, University of Science and Technology of China \& State Key Laboratory of Cognitive Intelligence, Hefei, China, \\ $^2$  Shenzhen International Graduate School, Tsinghua University, Shenzhen, China}
	\country{}}

\email{{mycheng, doge, prime666}@mail.ustc.edu.cn,{qiliuql, cheneh}@ustc.edu.cn, zhilizl@sz.tsinghua.edu.cn}

\thanks{Qi Liu is corresponding author. Our experiment codes are available at https://github.com/Mingyue-Cheng/FormerTime}

\renewcommand{\shortauthors}{Mingyue Cheng et al.}

\begin{abstract}
Deep learning-based algorithms, e.g., convolutional networks, have significantly facilitated multivariate time series classification (MTSC) task. Nevertheless, they suffer from the limitation in modeling long-range dependence due to the nature of convolution operations. Recent advancements have shown the potential of transformers to capture long-range dependence. However, it would incur severe issues, such as fixed scale representations, temporal-invariant and quadratic time complexity, with transformers directly applicable to the MTSC task because of the distinct properties of time series data. To tackle these issues, we propose FormerTime, an hierarchical representation model for improving the classification capacity for the MTSC task. In the proposed FormerTime, we employ a hierarchical network architecture to perform multi-scale feature maps. Besides, a novel transformer encoder is further designed, in which an efficient temporal reduction attention layer and a well-informed contextual positional encoding generating strategy are developed. To sum up, FormerTime exhibits three aspects of merits: (1) learning hierarchical multi-scale representations from time series data, (2) inheriting the strength of both transformers and convolutional networks, and (3) tacking the efficiency challenges incurred by the self-attention mechanism. Extensive experiments performed on $10$ publicly available datasets from UEA archive verify the superiorities of the FormerTime compared to previous competitive baselines. 
\end{abstract}




\maketitle
\section{Introduction}
Multivariate time series~\cite{zheng2014time,gaoreinforcement,hou2022multi} are ubiquitous for many web applications~\cite{yin2016forecasting,sezer2020financial}, which are sequences of events acquired longitudinally and each event is constituted by observations recorded over multiple attributes. For example, the electrocardiogram (ECG) signals~\cite{sarkar2020self} in electronic health records (EHRs) can be formulated as multivariate time series data since they can be obtained over time and multiple sensors. Comprehensive analysis of such data can facilitate decision-making in real applications~\cite{rim2020deep,Liu2022OnePO}, such as human activity recognition, healthcare monitoring, and industry detection. Particularly, multivariate time series classification (MTSC) tasks, as one fundamental problem of time series analysis, received significant attention in both academia and industry. 

Accordingly, numerous efforts~\cite{zhang2020tapnet,tan2022multirocket} have been devoted to the MTSC problem over the last decades. In general, most of these current works can be divided into two categories: pattern-based and feature-based models. The former type usually extracts useful bag-of-patterns or shapelet patterns from the whole time series, and then transforms these extracted patterns into features to be used as inputs for the classifier. Since they are generated in raw time series, the corresponding patterns are often interpretable. The main concern is the often-incurred expensive computation cost during the process of pattern extraction. In contrast, feature-based methods can be very efficient and can be scaled to large-scale time series data but their classification capacity greatly depends on the effectiveness of labor-intensive features based on domain experts.

Hence, researchers begin to explore more expressive feature maps for improving classification capacity. Deep learning-based models~\cite{zheng2014time,goel2017r2n2,ismail2019deep,karim2019multivariate,zhang2020tapnet,tran2021radflow,cheng2022towards} have achieved remarkable success and have become ever-increasingly prevalent over past advancements. The main reason is that discriminative features related to time series can be learned in an end-to-end manner, which significantly saves manual feature engineering efforts. Among them, convolutional-based methods nearly have become the dominant approach due to the strong representation capacity of convolution operations~\cite{ruiz2021great,cheng2022towards}. In general, the strengths of convolutional models in performing time series classification can be summarized as follows: (1) convolutional networks can easily learn multi-scale representations by controlling the strides of convolutional kernels ~\cite{wang2021pyramid}, and preserve the capacity of temporal-invariant via the weight-sharing mechanism, which are of great significance for MTSC tasks; (2) Very deep convolutional networks can be stacked by employing residual connections, enabling larger receptive fields for capturing the sequence dependence; (3) Convolution operations can be efficiently computed without suffering from limitations of sequence length or instance number, and thus can be easily scaled to massive datasets. Despite their effectiveness, we argue that the classification performance is still restricted in failing to capture global contexts in convolution operation.

Recently proposed transformer architecture~\cite{vaswani2017attention,Wu2022FlowformerLT} have shown promising capacity in capturing global contexts for language modeling tasks. Motivated by this, we seek to transfer the powerful capacity of transformers from language domain to time series. However, it would easily incur severe issues in applying the transformer to the MTSC problem. First, in language transformers, only fixed-scale representations can be learned since word tokens serve as the basic elements. By contrast, the information density of a single object in time series is too small to reflect helpful patterns related to class labels. Taking an example in ECG classification,  informative patterns are typically characterized by a series of continuous points or various sub-series instead of a single point. As such, multi-scale feature maps~\cite{tang2021omni} are necessary and can take a significant influence on the classification capacity. Second, the capacity of temporal-invariant is largely weakened in vanilla transformers since self-attention is permutation-variant, which may also restrict the final performance~\cite{oh2018learning}. Last, the sequence length of time series usually can be much longer than language sentences, which inevitably incurs an expensive computation burden since the time and memory complexity of the self-attention mechanism in the transformer architecture is quadratic to the sequence length input. Though some pioneering efforts~\cite{zerveas2021transformer,liu2021gated, tay2020efficient} based on transformers have been devoted before, the problems discussed above are still under exploration in the MTSC task.

To tackle these severe issues, in this work, we present FormerTime, a hierarchical transformer network for the MTSC task. Specifically, we design a hierarchical structure by dividing the whole network into several different stages, with different levels of scales as input. We also develop a novel transformer encoder to perform hidden transformation with two distinct characteristics: (1) we replace the standard self-attention mechanism with our newly designed temporal reduction version to save computation cost; (2) we design a context-aware positional encoding generator, which is designed to not only preserve the order of sequence input but also enhance the capacity of temporal-invariant of the whole model. In general, our FormerTime exhibits the following merits. First, in contrast to convolutional-based classifiers, FormerTime always yields a global reception field, which is useful for capturing the long-range dependence and interaction of the whole time series. Second, FormerTime can conveniently learn time series representations on various scales via its hierarchical architecture. Third, in the FormerTime, the inductive bias of temporal-invariant is well enhanced by leveraging the contextual positional generators. More importantly, the computation consumption of FormerTime is largely saved, and could be acceptable even for very long sequences.  

To evaluate the effectiveness of the FormerTime, we conduct extensive empirical studies on $10$ public datasets from the UEA archive~\cite{bagnall2018uea}. The experimental results clearly show that FormerTime can yield strong classification performance in average. It significantly outperforms compared strong baselines. In summary, we initially demonstrate the potential of transformers in the MTSC problem. To the best of our knowledge, we first the few attempts of transformer-based models in breaking the efficiency bottleneck and achieving great performance improvements. We hope our work can facilitate the study of applying transformers to the MTSC problem.

\vspace{-0.1in}
\section{Related Work}
\subsection{Time Series Classification} 
Tremendous efforts have been devoted to time series classification. Generally speaking, previous works can be roughly divided into two types: pattern-based and feature-based methods. Pattern-based methods typically first extract bag-of-patterns~\cite{schafer2017multivariate}  or shapelet patterns~\cite{ye2009time}, and then feed them into a classifier. For example, shapelet-based methods usually extract some useful subsequence, which is distinctive for different classes. Then, distance metrics are employed to generate features for classification. DTW has been proved as an effective distance measurement in time series classification. Recently proposed work~\cite{schafer2017multivariate} uses symbolic Fourier approximation to generate discrete units for classification. The strength of these methods is that the generative patterns are usually interpretable~\cite{crabbe2021explaining} while the largest weakness~\cite{bagnall2017great,zhang2020tapnet} is inevitablely incurring expensive computation in producing discriminative patter features. A series of methods have been proposed to overcome this weakness by either speeding up the computation of distance or constructing the dictionary~\cite{ye2009time}. In contrast to pattern-based methods, feature-based methods can be more efficient by depending on hand-crafted static features based on domain experts. However, it is difficult to design good features to capture intrinsic properties embedded in various time series data. Therefore, the accuracy of feature-based methods is usually worse than that of sequence distance-based ones. Recently, extracting time series feature with deep neural networks~\cite{zheng2014time,karim2019multivariate,ismail2019deep} gradually become prevalent in time series classification tasks. As a pioneering attempt, multi-channel convolutional networks~\cite{zheng2014time} have been proposed to deal to capacity of classification for the MTSC problem.  Later, a series of convolutional-based methods, like ResNet~\cite{he2016deep}, InceptionTime~\cite{ismail2020inceptiontime} have also proposed to achieve remarkable success due to their powerful representation ability in extracting time series features. ROCKET~\cite{dempster2020rocket} employs random convolution kernels to train linear classifiers and has been recognized as a powerful method in recent empirical studies~\cite{ruiz2021great}. Other than pure convolutional based methods, ~\cite{karim2019multivariate} perform time series classification by designing a hybrid network, in which both the LSTM layer and stacked convolutional layer along with a squeeze-and-excitation block are simultaneously used to extract features. Besides, mining graph structure among time series with graph convolutional networks for comprehensive analysis have also attracted some researchers~\cite{wu2020connecting,duan2022multivariate}.  
\subsection{Transformers in Time Series.} Designed for sequence modeling, the transformer network~\cite{vaswani2017attention} recently achieves great success in nature language processing (NLP). Among multiple advantages of transformers, the ability to capture long-range dependencies and interactions is especially attractive for time series modeling. Hence, a large body of transformer-based methods~\cite{zhou2021informer,wen2022transformers} has been proposed by attempting leveraging transformers for time series forecasting or regression. For time series classification, recent advancements still lie in the early stage and mainly focus on multivariate time series classification. \cite{russwurm2020self} studies the transformer for raw optical satellite time series classification and obtains the latest results comparing with convolution-based solutions. GTN~\cite{liu2021gated} explores an extension of the transformer network by modeling both channel-wise and step-wise correlations, simultaneously. Besides, pre-trained transformers are also investigated in time series classification tasks. For example, TST~\cite{zerveas2021transformer} employs the transformer network to learn unsupervised representations of time series so as to alleviate the data sparsity issue. Despite their effectiveness, both the efficiency issues incurred by the self-attention and the properties of time series classification task are largely ignored in these current works. Although some prior works focusing on improving the efficiency of self-attention were developed~\cite{tay2020efficient}, these works mainly use heuristic strategies to perform sparse attention computation without global context modeling. In this work, we aim to tackle the efficiency bottleneck and achieve performance improvements  in the proposed FormerTime. 
\begin{figure*}[t]
	\centering
	\includegraphics[width=1.0\textwidth]{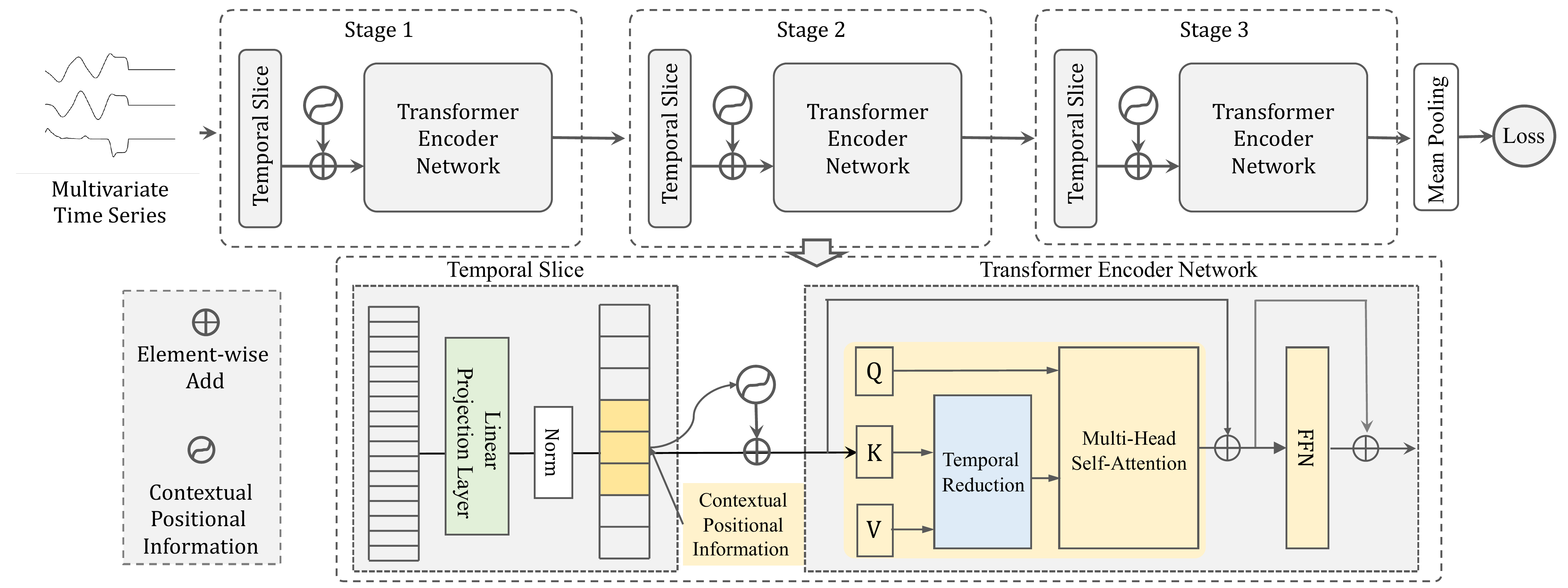} 
	\vspace{-0.2in}
	\caption{Illustration of the FormerTime, i.e., a efficient hierarchical transformer architecture for the MTSC task.}
		\vspace{-0.12in}
	\label{fig:formertime}
\end{figure*}
\section{The FormerTime Model}
In this section, we first formally introduce the definition of multivariate time series classification (MTSC) problem. Then, we introduce the overall architecture of our designed FormerTime. After that, we elaborate the FormerTime via respectively introducing two aspects of key designs, i.e., the hierarchical architecture and the designed transformer encoder. Finally, we demonstrate the difference between the FormerTime and other relative methods.  
\subsection{Problem Definitions}
We introduce the definitions and notations used in the following. $\mathbb{D} = {(X^1, y^1), (X^2, y^2),..., (X^n,y^n)}$ is a dataset containing a collection of pairs $(X^i, y^i)$, in which $n$ denotes the number of examples and $X^i$ denote a multivariate time series with its corresponding label denoted by $y^i$ . Each multivariate time series $X = [x_1, x_2, ..., x_l]$ contains $l$ ordered elements with $m$ dimensions in each time step. The task of multivariate time series classification is to learn a classifier on $\mathbb{D}$ so as to map from the space of inputs $X$ to a probability distribution over the class $y$. 
\subsection{Model Architecture Overview}
An overview of the FormerTime is depicted in Figure~\ref{fig:formertime}. The input of the whole model is a set of multivariate time series, involving multiple dimensions (a.k.a. channels). For each dimension, the time series share the same sequence length. To produce multi-scale representations for time series data, we adopt a hierarchical architecture. Specifically, we divide the whole deep network architecture into multiple different stages so as to generate feature maps on various time scales. For simplicity, all stages share a similar architecture, which is composed of a temporal slice partition processing operation and successive $L_i$ our designed transformer encoder layers. Relying upon such hierarchical architecture, the time series representation on various scales can be effectively extracted. Meanwhile, we use the mean pooling operation over the representation of each temporal point to denote the time series representations. The model's output is the predicted distribution of each possible label class for the given input time series. The FormerTime model can be trained end-to-end by minimizing the cross-entropy loss~\cite{Cheng2021LearningRS}, and the loss is propagated back from the prediction output across the entire network. The following sections would mainly introduce several key designs in the FormerTime. 
\subsection{Multi-scale Time Series Representations}
Integrating information on different time scales~\cite{chen2021multi} is essential to the classification capacity in the MTSC task. However, the vanilla transformer model only can produce fixed-scale representations of the sequence input. Thus, we devise a hierarchical architecture for learning time series representations on various time scales. The key idea is that the whole model is divided into several stages so as to hierarchically performing feature maps. Here, we vary the time scale input for different stages by leveraging a temporal slice partition strategy, i.e., aggregating successive neighborhood points with a window slicing operation. 

Specifically, suppose that we have sequence input $X=\{x_1, x_2, ...x_l\}$  in stage $j$, and the window slicing size is $s_j$, which denotes the scale size of the processed time series. Every $s_j$ successive points will be grouped into a new temporal slice. Considering the semantic gap issue, we then feed these temporal slices to a trainable linear projection layer to project this raw feature to a new dimension $C_j$. Notably,  the weights of this linear projection layer is weight-shared, in which we use $d_j$ to denote the stride of projection operations. To some extent, these temporal slices can be regarded as the ``tokens'' of new time series, analogous to the relationship between word and whole speech input. Assume that the whole model is divided into three stages, and the fine-grained raw time series can be processed into a new granularity version, which contains $\frac{l}{s_1}$ slices and each size is $s_1\times m$. Then, the linear projection layer project it into a new dimension $C_1$, and the output is reshaped to size of $\textbf{F}_1\in \mathbb{R}^ {\frac{l}{s_1}\times C_1} $. After that, the normalized embeddings of each temporal slice along with its positional embeddings are fed into the transformer encoder with $L_1$ layers. In the same manner, by using the feature maps of the previous stage's output as the input of the next stage, different scales of time series representations, denoted by $\textbf{F}_j$, can be effectively learned by stage-wise propagation. 

The primary motivation is the information density difference between time series and language domain. Unlike the tokens in language data, which are human-generated signals and highly semantic and information-intensive, the time series are naturally redundant, e.g., a missing point can be easily recovered from its neighbors. The benefits of adopting temporal slice operation are two-fold. First, the time scales of sequence data can be flexibly transformed, naturally forcing the network to generate hierarchical feature maps. Second, with this partitioning strategy, the sequence length of the whole time series can be largely reduced before sent into the encoder, saving an amount of computation consumption. 
\subsection{The Transformer Encoder Network}
In this subsection, we will introduce our designed encoder to extract the robust global contexts of the whole sequence input.
\subsubsection{Temporal Reduction Attention.}
To capture the global contexts of the whole time series, we would like to benefit from regarding the transformer network as the encoder to perform non-linear hidden representation transformation. One of the core designs of the vanilla transformer network is to employ a multi-head self-attention mechanism. Each temporal point needs to be computed the attention scores among all other sequence points with the inner product so as to capture the long-range dependence. However, the computation complexity incurred by the attention operation can be very heavy, growing quadratically in the sequence length of the input sequence. Unlike language data, the sequence length of time series data regularly can be very long to uncover the event of classification tasks. Directly leveraging the standard attention computation strategy is memory-expensive, making it hard for transformers to be applicable in time series.

To solve this dilemma, inspired by recent works~\cite{wang2020linformer}, we present a novel attention computation strategy named temporal reduction attention  (TRA) mechanism to replace the vanilla self-attention strategy. The core idea of TRA is to compute the attention with a sub-sampled version of all input points. To be more specific, similar to standard self-attention, our TRA receives a query $Q$, a key $K$, and a value $V$ as input, and outputs a refined feature. Details of TRA operation in stage $j$ can be formulated as follows:
\begin{small}
	\begin{equation}
		\label{equ:tra_qkv}
		\rm TRA (\textbf{Q}, \textbf{K}, \textbf{V}) = Concat(head_0, ..., head_{N_j})\textbf{W}^O,
	\end{equation}
\end{small}\noindent in which the $\rm Concat$ denotes the concatenation operation, and $N_i$ is the number of heads in the attention layer.
\begin{small}
	\begin{equation}
		\label{equ:head}
		\rm head_j =Attention(\textbf{Q}\textbf{W}_j^Q, TR(\textbf{K}) \textbf{W}_j^K, TR(\textbf{V})\textbf{W}_j^V),
	\end{equation}
\end{small}\noindent where $\rm \textbf{W}_j^Q\in \mathbb{R}^{C_j\times d_{head}}, \textbf{W}_j^K\in \mathbb{R}^{\textbf{C}_j\times d_{head}}, \textbf{W}_j^V\in \mathbb{R}^{\textbf{C}_j\times d_{head}}$ are linear projection parameters. In this way, the dimension of each head is equal to $\frac{\textbf{C}_j}{N_j}$. Here, $\rm TR$ indicates the temporal reduction on $K$ and $V$, which can be written as
\begin{small}
	\begin{equation}
		\label{equ:tr}
		\rm TR(x) = Norm(Reshape(x, R_j)\textbf{W}^T), 
	\end{equation}
\end{small}\noindent where $\textbf{x}\in\mathbb{R}^{l_j \times C_j}$ represents a input sequence, and $R_i$ denotes the reduction ratio of the attention layers in stage $j$.  $\rm Reshape(x, R_j)$ is an operation of reshaping the input sequence $x$ to a sequence of size $\frac{L_j}{R_j}$ and $\textbf{W}^T$  is a linear projection that reduces the dimension of the input sequence to $C_j$. Here, we employ layer normalization operation~\cite{vaswani2017attention} to implement $\rm Norm$. Like the original attention computation mechanism, our $\rm Attention(\cdot)$  can be represented as 
\begin{small}
	\begin{equation}
		\label{equ:attention}
		\rm Attention(q,k, v) = softmax(\dfrac{qk^\intercal}{\sqrt{d_{head}}})v.
	\end{equation}
\end{small}\indent
In this way, the computational cost of our attention operation is $\frac{1}{R_i}$ of standard self-attention, so our TRA can handle longer input feature maps/sequences without requiring too many resources. The main difference between our TRA and the prior version is that our TRA reduces the temporal scale of $\textbf{K}$ and value $\textbf{V}$ before the attention operation, largely reducing the computational/memory overhead. Note that the capacity of global context modeling is still well-remained in our attention layer.
\subsubsection{Contextual Positional Encoding.} As claimed in~\cite{vaswani2017attention,chu2021conditional}, positional information is a key operation in the success of the transformer network. The main reason is that the self-attention operation can be used to preserve the order of the sequence property of input data. Two types of position information in transformers have been widely adopted, including absolute and relative encodings, which respectively denote static and learnable embeddings. For the former type, absolute temporal information provides helpful cues for whenever the object would appear in the whole time series. Despite its effectiveness, it severely weakens the capacity of temporal-invariant since each temporal slice is added with a unique positional encoding. In fact, the temporal invariance plays a significant role in time series classification tasks since we hope the model to release the same response whenever the discriminative pattern appears in the time series. Though relative positional encodings can greatly alleviate the aforementioned issues, the relative encodings lack absolute position information, which is important in classification tasks~\cite{islam2020much,chu2021conditional}. Previous works have uncovered that one of the main merits of such positional information is that absolute information can be added for enhancing classification performance. Besides, we also argue that these two types of positional information actually model each temporal slice individually and only achieve sub-optimal performance because that extracted patterns from time series evolve in time and highly depend on their surrounding points. 

Based on the above analysis, we hold that the well-informed positional information should  possess two aspects of characteristics. First, making the input sequence permutation-variant but temporal-invariant is a necessity for time series classification. Second, having the ability to provide absolute information also matters. To fulfill the two demands, we find that characterizing the contextual information among neighboring temporal slices can be sufficient. As shown in Figure~\ref{fig:formertime}, we use $1$-D convolutional kernel size $k$ along with $\frac{k}{2}$ zero paddings to extract the localized contextual information as positional encodings. Note that the zero padding is vital to make the model aware of the absolute position information.  
\subsubsection{Entire Transformer Encoder Network.}
Based on the designed temporal reduction attention mechanism and context positional encodings, we organize them together to form a novel transformer encoder block for learning the time series representations. We give a sketch of the newly designed encoder at the bottom of Figure~\ref{fig:formertime}. On the basis of the design of standard transformer encoder~\cite{vaswani2017attention}, the entire encoder is composed of successive layers of the TRA layer and followed by a feed-forward neural network (FFN) layer. These two layers are further wrapped with a residual connection to avoid the the vanishing gradient problem. Particularly, we set a trainable parameter $\alpha$, initialized with zero, according to the previous work~\cite{bachlechner2021rezero}. Such a simple trick can further help the FormerTime converge more stable.
\subsection{Summary and Remarks}
In the following, we summarize the characteristics of FormerTime and discuss its relations to transformer-based and convolutional-based approaches in the multivariate time series classification.
\paragraph{\textit{Relation to Transformer-based Models}.}
Transformer architecture has shown its superiority in global sequence modeling tasks. However, the dot-product computation in self-attention easily incurs quadratic computation and memory consumption on sequence input. Hence, the efficiency of the transformer architecture becomes the bottleneck of applying them to time series classification tasks. Besides, the vanilla transformer model lacks some key designs of inductive bias, such as multi-scale representation and temporal invariance, which can greatly benefit the time series classification task. Although some prior works of efficient transformer variants~\cite{tay2020efficient} have been proposed, they fail to solve these two aspects of limitations, simultaneously. In contrast, the proposed FormerTime not only breaks the bottleneck of efficiency but also achieves performance improvements in the MTSC task.
\paragraph{\textit{Relation to Convolutional-based Models}.}
Recently, researchers have demonstrated the extremely expressive capacity of convolutional models in MTSC tasks. In fact, convolutional networks can yield several aspects of strength: 1) their memory and time complexity in feature extraction are not constrained by the sequence length of the sequence input, 2) they can easily learn time series with various scales of feature maps with varying strides and preserve the prior capacity of temporal-invariant with the weight-sharing mechanism for the classification task. However, convolution operation cannot achieve a global receptive field of whole sequence input while the global context information is vital for the classification capacity. Our proposed FormerTime not only absorbs the strength of convolutional models but also meets the demands of long-range dependence modeling. 
\begin{table}[t]
	\centering
	\setlength{\tabcolsep}{1.0pt}
		\vspace{-0.12in}
	\caption{Statics of datasets in the experiments. }
	\vspace{-0.1in}
	\begin{tabular}{cccccc}
		\hline
		Dataset  & Train Size & Test Size & Dimensions & Length & Classes \\
		\hline
		AWR   & 275   & 300   & 9     & 144   & 25 \\
		AF    & 15    & 15    & 2     & 640   & 3 \\
		CT    & 1,422  & 1,436  & 3     & 182   & 20 \\
		CR    & 108   & 72    & 6     & 1,197  & 12 \\
		FD    & 5,890  & 3,524  & 144   & 62    & 2 \\
		FM    & 316   & 100   & 28    & 50    & 2 \\
		MI    & 278   & 100   & 64    & 3,000  & 2 \\
		SRS1  & 268   & 293   & 6     & 896   & 2 \\
		SRS2  & 200   & 180   & 7     & 1,152  & 2 \\
		UWG    & 120   & 320   & 3     & 315   & 8 \\
		\hline
	\end{tabular}%
	\vspace{-0.2in}
	\label{tab:datasets}%
\end{table}%
\vspace{-0.1in}
\section{Experiments}
\subsection{Experimental Setup}
\subsubsection{Datasets} 
We perform all experiments by conducting experiments on ten public datasets, which are selected from the well-known UEA multivariate time series classification (MTSC) archive. In reality, the UEA archive has become nearly the most widely used multivariate time series benchmarks. We select a set of 10 multivariate datasets from the UEA archive~\cite{bagnall2018uea} with diverse characteristics in terms of the number, and the length of time series samples, as well as the number of classes. Specifically, we choose: ArticularyWordRecognition (AWR), Atrial Fibrillation (AF), CharacterTrajectories (CT), Cricket, FaceDetection (FD), FingerMovements (FM), MotorImagery(MI), SelfRegulationSCP1 (SRS1), SelfRegulationSCP2 (SRS2), UWaveGestureLibrary (UW). In these original dataset, training and testing set have been well processed. We do not take any processing for these datasets for a fair comparison. We summarize the main characteristics of dataset in Table~\ref{tab:datasets}.

\begin{table*}
	\centering
	\setlength{\tabcolsep}{3.0pt}
	\caption{Detailed Hyper-parameter settings of the proposed FormerTime. }
		\vspace{-0.1in}
	\begin{tabular}{c|cc|cc|cc}
		\hline
		\multirow{2}[2]{*}{Datasets} & \multicolumn{2}{c|}{Stage 1} & \multicolumn{2}{c|}{Stage 2} & \multicolumn{2}{c}{Stage 3} \\
		& Temporal Slice &  Encoder & Temporal Slice &  Encoder & Temporal Slice &  Encoder \\
		\hline
		AWR   & s\_1=2,C\_1=64,d\_1=2 & \multirow{10}[2]{*}{\makecell{L\_1=6 \\ R\_1=2 \\ N\_1=4 }} & s\_2=2,C\_2=64,d\_2=2 & \multirow{10}[2]{*}{\makecell{L\_2=6 \\ R\_2=2 \\ N\_2=4}} & s\_3=2,C\_3=64,d\_3=2 & \multirow{10}[2]{*}{\makecell{L\_3=6 \\ R\_3=1 \\ N\_3=4} } \\

		AF    & s\_1=16,C\_1=64,d\_1=8 &       & s\_2=2,C\_2=64,d\_2=2 &       & s\_3=2,C\_3=64,d\_3=2 &  \\
		CT    & s\_1=16,C\_1=64,d\_1=8 &       & s\_2=2,C\_2=64,d\_2=2 &       & s\_3=2,C\_3=64,d\_3=2 &  \\
		CR    & s\_1=8,C\_1=64,d\_1=8 &       & s\_2=2,C\_2=64,d\_2=2 &       & s\_3=2,C\_3=64,d\_3=2 &  \\
		FD    & s\_1=2,C\_1=64,d\_1=2 &       & s\_2=2,C\_2=64,d\_2=2 &       & s\_3=2,C\_3=64,d\_3=2 &  \\
		FM    & s\_1=4,C\_1=64,d\_1=4 &       & s\_2=2,C\_2=64,d\_2=2 &       & s\_3=2,C\_3=64,d\_3=2 &  \\
		MI    & s\_1=8,C\_1=64,d\_1=8 &       & s\_2=2,C\_2=64,d\_2=2 &       & s\_3=2,C\_3=64,d\_3=2 &  \\
		SRS1  & s\_1=2,C\_1=64,d\_1=2 &       & s\_2=2,C\_2=64,d\_2=2 &       & s\_3=2,C\_3=64,d\_3=2 &  \\
		SRS2  & s\_1=16,C\_1=64,d\_1=8 &       & s\_2=2,C\_2=64,d\_2=2 &       & s\_3=2,C\_3=64,d\_3=2 &  \\
		UWG   & s\_1=8,C\_1=64,d\_1=8 &       &s\_2=2,C\_2=64,d\_2=2&       & s\_3=2,C\_3=64,d\_3=2 &  \\
		\hline
	\end{tabular}%
  	\vspace{-0.1in}
	\label{tab:formertime}%
\end{table*}%

\begin{table*}[htbp]
  \centering
  \caption{Classification performance of compared methods in ten datasets. \textbf{Bold} numbers represent the best results.}
  	\vspace{-0.1in}
    \begin{tabular}{ccccccccccccc}
    \toprule
    Datasets & IT    & LS    & ST    & MCDCNN & TCN   & MCNN  & ResNet & MR    & TST   & GTN   & Informer & Ours \\
    \midrule
    AWR   & 0.9827  & 0.9127  & 0.8700  & 0.7800  & 0.9467  & 0.8200  & 0.9827  & 0.9720  & 0.9789  & 0.9767  & 0.9820  & \textbf{0.9847 } \\
    AF    & 0.4400  & 0.2533  & 0.2667  & 0.3733  & 0.4933  & 0.3467  & 0.4000  & 0.3333  & 0.4000  & 0.4000  & 0.4267  & \textbf{0.6000 } \\
    CT    & \textbf{0.9983 } & 0.9866  & 0.7224  & 0.8826  & 0.9915  & 0.9238  & 0.9965  & 0.9876  & 0.9882  & 0.9783  & 0.9862  & 0.9914  \\
    CR    & 0.9889  & 0.9639  & 0.9722  & 0.6278  & 0.9083  & 0.9167  & \textbf{0.9972 } & 0.9806  & 0.9583  & 0.7917  & 0.9778  & 0.9806  \\
    FD    & 0.6820  & 0.5129  & 0.5085  & 0.5000  & 0.6801  & 0.6747  & 0.5760  & 0.6065  & 0.6005  & 0.5542  & 0.5265  & \textbf{0.6872 } \\
    FM    & 0.6000  & 0.4840  & 0.4940  & 0.5920  & 0.5880  & 0.5920  & 0.6080  & \textbf{0.6380 } & 0.5900  & 0.5350  & 0.6120  & 0.6180  \\
    MI    & 0.5860  & 0.5180  & 0.6100  & 0.5000  & 0.6040  & 0.5980  & 0.5780  & 0.5640  & N/A    & N/A    & 0.6240  & \textbf{0.6320 } \\
    SRS1  & 0.8942  & 0.7038  & 0.6724  & 0.9079  & 0.9031  & 0.8949  & 0.8730  & \textbf{0.9352 } & 0.8771  & 0.8019  & 0.9188  & 0.8867  \\
    SRS2  & 0.5689  & 0.5111  & 0.5300  & 0.5256  & 0.5978  & \textbf{0.5989 } & 0.5622  & 0.5411  & 0.5796  & 0.5611  & 0.5767  & 0.5922  \\
    UWG   & 0.8869  & 0.8031  & 0.7769  & 0.8438  & 0.7981  & 0.8044  & 0.7994  & \textbf{0.9075 } & 0.8271  & 0.8406  & 0.8363  & 0.8881  \\
    \hline
    Average & 0.7628  & 0.6649  & 0.6423  & 0.6533  & 0.7511  & 0.7170  & 0.7373  & 0.7466  & 0.7555  & 0.7155  & 0.7467  & \textbf{0.7861 } \\
    MACs (M) & 89    & -     & -     & 263   & 283   & 929   & 132   & -    & 408   & 1,565  & 141   & 98  \\
    \bottomrule
    \end{tabular}%
  	\vspace{-0.1in}
  \label{tab:main_results}%
\end{table*}%

\subsubsection{Compared Baselines.} 
For comprehensive evaluation, we choose the following prevalent baseline methods for evaluation: Shapelet Transformation (ST)~\cite{lines2012shapelet}, Learning Shapelet (LS)~\cite{grabocka2014learning}, TST~\cite{zerveas2021transformer}, GTN~\cite{liu2021gated}, Informer~\cite{zhou2021informer}, MCDCNN~\cite{zheng2014time}, MCNN~\cite{cui2016multi}, ResNet~\cite{he2016deep}, TCN~\cite{bai2018empirical}, InceptionTime (IT)~\cite{ismail2020inceptiontime}, MiniROCKET (MR)~\cite{dempster2021minirocket}. Among them, ST and LS are two shapelet-based methods. TST and GTN are two transformer-based models proposed for time series classification. Though Informer is originally proposed for time series forecasting, we also treat it as a competitive baseline to verify the effectiveness of our FormerTime. In addition, the remaining compared baselines are convolutional-based models applied to the MTSC problem.  Note that some traditional classifiers~\cite{middlehurst2021hive} are not considered here, since it is difficult to construct hand-crafted features for all time series. Also, we do not choose well-known distance-based methods, like HIVE-COTE~\cite{middlehurst2021hive}, as baseline due to their expensive computation consumption. We adopt accuracy as the metric. 
\subsubsection{Implement Details.}
For learning shapelet (LS), we adopt the codes in \footnote{https://tslearn.readthedocs.io/} while adopting the publicly available codes~\footnote{https://pyts.readthedocs.io/} to run shapelet transformation (ST). To implement GTN, we use the source code provided by the corresponding authors \footnote{https://github.com/ZZUFaceBookDL/GTN}. We implement TST by strictly following the network architecture settings of original works using PyTorch. We replace the decoder in Informer\footnote{https://github.com/zhouhaoyi/Informer2020} with a linear classifier layer so as to adapt it for MTSC tasks. The remaining baselines' code consistently leverages the codes in~\footnote{https://timeseriesai.github.io/tsai/}. For full reproducibility of the experiments, we release our codes and make it available ~\footnote{https://anonymous.4open.science/r/FormerTime-A17E/}. The specific hyper-parameters of our FormerTime are listed as follows:
\begin{itemize} 
	\item $s_j$: the temporal slice size of stage $j$;
	\item $C_j$: the hidden size of the output in stage $j$;
	\item $d_j$: the stride size of window slicing operation of stage $j$;
	\item $L_j$: the number of transformer encoders in stage $j$;
	\item $R_j$: the temporal reduction rate in stage $j$;
	\item $N_j$: the number of attention heads of temporal reduction attention in stage $j$. 
\end{itemize}
More details of hyper-parameter settings in FormerTime for specific datasets can be found in Table~\ref{tab:formertime}. For common hyper-parameters of all models, we set the embedding size as $64$. The initialized learning rate is set to $1\times 10^{-3}$ without additional processing, and we employ Adam optimizer to guide all model training. All other hyper-parameters and initialization strategies either follow the suggestions from the original works' authors or are tuned on testing datasets. We report the results of each baseline under its optimal hyper-parameter settings. For a fair comparison, all models are trained on the training set and report the accuracy score on the testing set. All models are trained until achieving the best results. \textbf{All experiments in our work are repeated for $5$ times with $5$ different seeds, and we reported the mean value score.}
\begin{figure}
	\centering
	\includegraphics[width=0.46\textwidth]{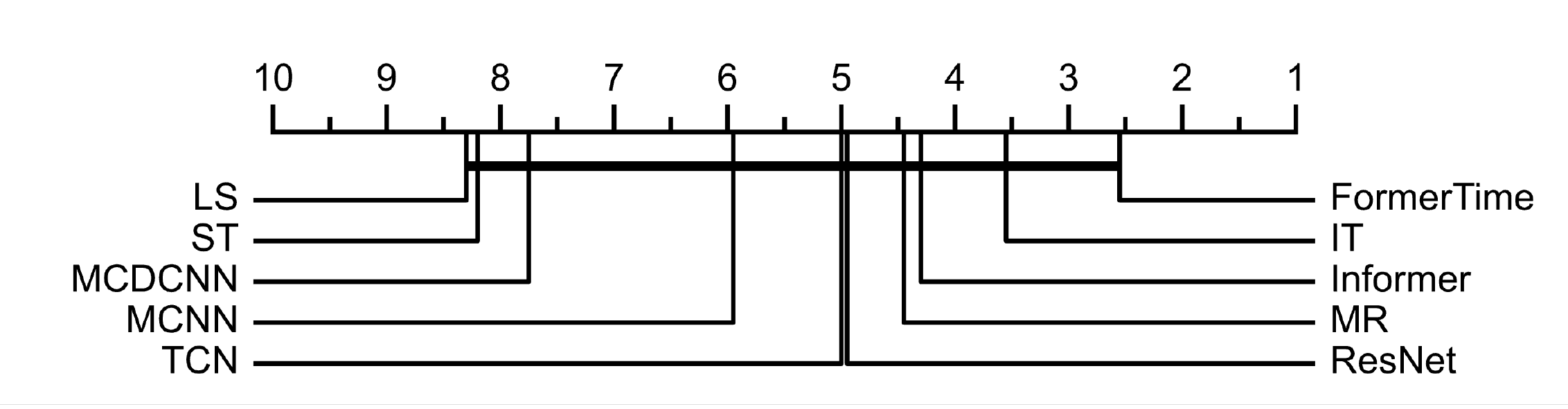} 
	\vspace{-0.1in}
	\caption{Critical difference diagram over the mean ranks of FormerTime, baseline methods. }
	\vspace{-0.1in}
	\label{fig:cdg}
\end{figure}
\subsection{Experimental Results}
\subsubsection{Classification Performance Evaluation.}
Table~\ref{tab:main_results} summarizes the classification accuracy of all compared methods while Figure~\ref{fig:cdg} reports the critical difference diagram as presented in~\cite{demvsar2006statistical}. The results of ``N/A'' indicates that the corresponding results cannot be run due to the out-of-memory issue. Overall, the accuracy of our proposed FormerTime could outperform previous classifiers on average. Such results demonstrate the success of FormerTime in enhancing the classification capacity in the MTSC problem.  For each dataset, the classification performance of FormerTime is either the most accurate one or very close to the best one. These existing proposed models typically cannot always achieve the most distinct results. One may wonder whether the FormerTime can be effective enough. However, the experimental results are largely consistent with previous empirical studies~\cite{bagnall2017great,ruiz2021great}, i.e., one single model cannot always achieve superior performances in all scenarios. In particular, we observe that FormerTime could surpass other baselines to a large margin in datasets of AF and MI, in which the sequence length of these two datasets is very long. We guess that our temporal slice setting can be very robust for these two datasets.
\begin{table}
	\centering
	\caption{Experimental results w.r.t. studying the hyper-parameter sensitivity with varying stages.}
	\vspace{-0.1in}
	\begin{tabular}{ccccc}
		\hline
		Datasets & 1     & 2     & 3     & 4 \\
		\hline
		AWR   & \textbf{0.9811}  & \textbf{0.9811}  & 0.9720  & 0.9767  \\
		AF    & 0.4222  & 0.4667  & \textbf{0.6000}  & 0.5778  \\
		CT    & 0.9907  & 0.9909  & \textbf{0.9914}  & 0.9902  \\
		CR    & \textbf{0.9861}  & 0.9815  & 0.9806  & 0.9769  \\
		FD    & 0.6750  & \textbf{0.6793}  & 0.6776  & 0.6748  \\
		FM    & \textbf{0.6200}  & 0.6033  & 0.6140  & 0.6067  \\
		MI    & 0.6200  & 0.6267  & \textbf{0.6280}  & 0.6133  \\
		SRS1  & 0.8760  & 0.8692  & 0.8771  & \textbf{0.8840}  \\
		SRS2  & 0.5722  & 0.5815  & \textbf{0.5922}  & 0.5889  \\
		UWG   & \textbf{0.9021}  & 0.8948  & 0.8844  & 0.8844  \\
		\hline
		Averge & 0.7645  & 0.7675  & \textbf{0.7817}  & 0.7774  \\
		\hline
	\end{tabular}%
	\vspace{-0.1in}
	\label{tab:stage}%
\end{table}%

For these baseline approaches, we observe that convolutional-based methods, like MR, exhibit strong classification performances in some datasets, which is analogous to the experimental results of recent empirical studies~\cite{ruiz2021great}. We hold the characteristics of multi-scale representation and temporal invariance of the convolution operations make a great contribution. Besides, in MR, the feature of PPV, denoting the proportion of positive values of extracted deep representations, also matters. However, for transformer-based classifiers, it seems that the performance cannot always outperform convolutional algorithms. We guess the main reason behind the performance is that: (1) the plain transformer architecture fails to learn hierarchical feature maps from time series data, and (2) the naive positional information might not be suitable for modeling time series since the semantic information of one single temporal point individually modeled.  Besides, compared to these deep learning-based methods, shapelet-based methods exhibit the worst classification performance due to a lack of poor representation capacity. However, in shapelet-based approaches, interpretable sub-sequence patterns can be extracted to make the model more understandable, which is vital in some applications. 

In time series classification tasks, model efficiency has always been an important concern. Here, we also show the computation cost by recording MACs\footnote{https://github.com/Lyken17/pytorch-OpCounter} of compared methods. Note that only methods trained with end-to-end manner are reported. Though we adopt a self-attention operation, the computation cost of our methods can be very economical. Particularly, compared to the standard transformer network, our proposed FormerTime could significantly save computation costs. The main reason could be attributed to two aspects: 1) we model the raw time series with hierarchical architecture, which significantly shorten the sequence length, and 2) we develop a temporal reduction layer to ensure each input point can attend to all other data points. 


\subsubsection{Study of Multi-scale Representations.} 
In this part, we decide to study the effectiveness of hierarchical feature maps in FormerTime by setting different types of model variants w.r.t. the different number of stages. Specifically, we report the average classification experimental results of ten datasets in Table~\ref{tab:stage}, varying the number of stages from 1 to 4. Note that the total number of layers is consistent in these model variants to eliminate the other influence factors. From the reported results, we observe that feature maps at various scales can indeed perform much better than the single-scale representation versions. Such results demonstrate the importance of multi-scale representation in time series classification tasks.  Furthermore, we also empirically analyze the effectiveness of hierarchical structure by performing hyper-parameter sensitivity analysis on the size of window slicing in temporal slice partition. The average accuracy of ten datasets is recorded in Table~\ref{tab:slice}. An attractive experimental phenomenon is that FormerTime equipped with a large slice size yields more promising results. We guess that the semantic information of a smaller temporal slice is too small to characterize discriminative patterns of distinguishing other examples. 

\begin{table}
	\centering
	\caption{Experimental results w.r.t. studying the hyper-parameter sensitivity w.r.t. temporal slice size.}
		\vspace{-0.1in}
	\begin{tabular}{ccccc}
		\hline
		Datasets & [16,32,64] & [8,16,32] & [4,8,16] & [2,4,8] \\
		\hline
		AWR   & 0.9720  & 0.9740  & 0.9820  & \textbf{0.9847}  \\
		AF    & \textbf{0.6000}  & 0.5600  & 0.4267  & 0.4400  \\
		CT    & \textbf{0.9914}  & 0.9886  & 0.9868  & 0.9873  \\
		CR    & \textbf{0.9806}  & 0.9806  & 0.9778  & 0.9667  \\
		FD    & 0.6776  & \textbf{0.6794}  & 0.6823  & 0.6872  \\
		FM    & 0.6140  & 0.6080  & \textbf{0.6180}  & 0.6040  \\
		MI    & \textbf{0.6280}  & \textbf{0.6280}  & 0.6160  & 0.6180  \\
		SRS1  & 0.8771  & 0.8826  & 0.8710  & \textbf{0.8867 } \\
		SRS2  & \textbf{0.5922}  & 0.5811  & 0.5856  & 0.5600  \\
		UWG   & 0.8844  & \textbf{0.8881}  & 0.8781  & 0.8775  \\
		\hline
		Averge & \textbf{0.7817}  & 0.7770  & 0.7624  & 0.7612  \\
		\hline
	\end{tabular}%
	\label{tab:slice}%
\end{table}%

\begin{table}
	\centering
				\vspace{-0.12in}
	\caption{Experimental results w.r.t. studying the effectiveness of contextual positional embeddings.}
			\vspace{-0.12in}
	\begin{tabular}{ccccc}
		\hline
		Datasets & None& Static & Learnable  & Ours \\
		\hline
		AWR   & 0.9433  & 0.9822  & 0.9811  & \textbf{0.9720}  \\
		AF    & 0.4667  & 0.5111  & 0.5556  & \textbf{0.6000}  \\
		CT    & 0.9821  & 0.9902  & 0.9863  & \textbf{0.9914}  \\
		CR    & \textbf{0.9815}  & 0.9676  & 0.9769  & 0.9806  \\
		FD    & 0.6740  & \textbf{0.6804}  & 0.6774  & 0.6776  \\
		FM    & 0.5900  & 0.5867  & \textbf{0.6200}  & 0.6140  \\
		MI    & 0.6233  & 0.5833  & 0.6167  & \textbf{0.6280}  \\
		SRS1  & 0.8635  & \textbf{0.8817}  & 0.8749  & 0.8771  \\
		SRS2  & 0.5704  & 0.5759  & \textbf{0.6018}  & 0.5922  \\
		UWG   & 0.8479  & 0.8729  & 0.8677  & \textbf{0.8844}  \\
		\hline
		Averge & 0.7543  & 0.7632  & 0.7758  & \textbf{0.7817}  \\
		\hline
	\end{tabular}%
	\label{tab:pos}%
	\vspace{-0.2in}
\end{table}%

\begin{figure*}[t]
	\centering
	\includegraphics[width=1.0\textwidth]{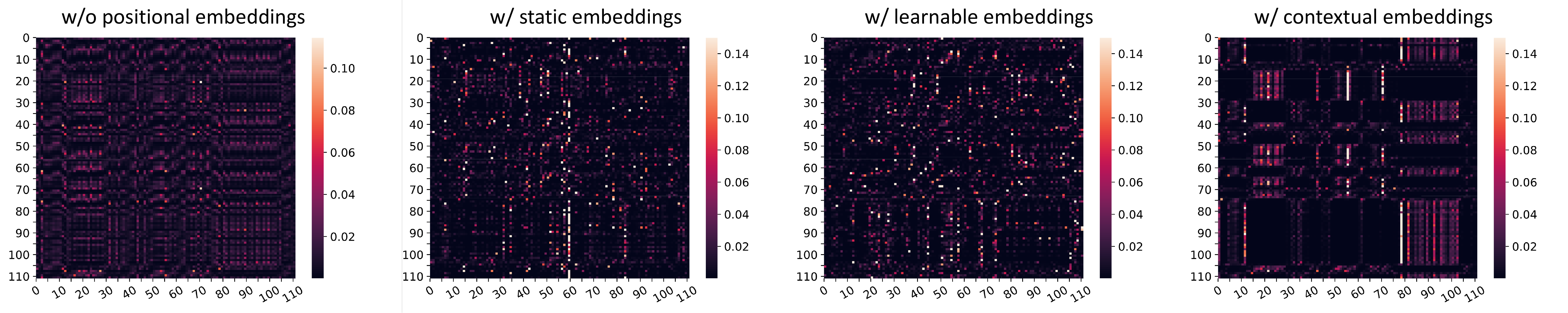} 
		\vspace{-0.2in}
	\caption{Normalized attention score from the first encoder block of the first stage in FormerTime: (1) without taking positional information into account, (2) using static embeddings, (3) using learnable vectors, (4) using our contextual embeddings.}
		\vspace{-0.1in}
	\label{fig:position_attention_score}
\end{figure*}
\begin{figure}[t]
	\centering
	\includegraphics[width=0.5\textwidth]{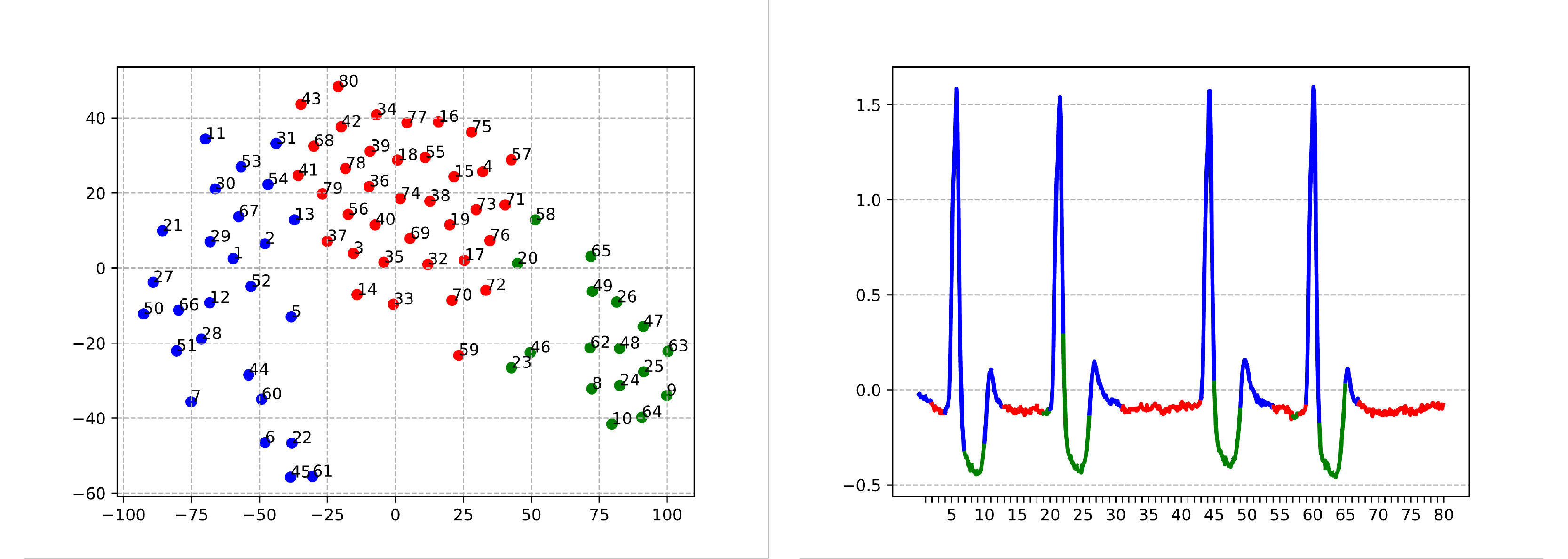} 
		\vspace{-0.2in}
	\caption{Left plot: Visualization of the t-SNE result of the embedding layer output on the AF dataset. Right plot: visualization of sub-sequences on raw time series data.}
		\vspace{-0.2in}
	\label{fig:semantic}
\end{figure} 
\subsubsection{Study of Positional Information Encoding.} In the following, we would like to empirically verify the effectiveness of contextual positional information. A natural choice is to replace the contextual embedding with static or learnable version~\cite{vaswani2017attention}, respectively. Moreover, we also evaluate the results of FormerTime without leveraging any forms of positional embedding information. The average experimental results conducted on ten datasets are shown in Table~\ref{tab:pos}. From these results, we notice that  FormerTime equipped with contextual positional information could surpass all other model variants, verifying the effectiveness of extracting contextual information as positional encodings. Also, FormerTime's performance dramatically degrades while absolutely discarding the positional information. We believe this is reasonable because self-attention computation is permutation-variant, whose performance would dramatically degrade if discarding the positional information. To further deeply understand the scheme of several types of positional encoding, we choose one sample from SRS1 data to visualize the attention weights of corresponding model variants. As shown in Figure~\ref{fig:position_attention_score}, unlike the widely adopted static and learnable positional embeddings, more specific attention map patterns could be well learned by our contextual positional information generating strategies. Also, it seems that most of the temporal slices would produce uniform attention weights if we remove the positional information. We believe these visualized cases further demonstrate the effectiveness of our contextual information-generating strategies.

\subsubsection{Analyzing Semantic of Time Series Slice Representations}
In this part, we aim to analyze the semantic descriptions of constructed temporal slices encoded by the embedding layer. To achieve this goal, we apply t-SNE to reduce the embedding of each temporal slice on an example selected from AF datasets. For better understanding, we further map each temporal slice to the raw time series. As shown in Figure~\ref{fig:semantic}, we find that the similarity of raw time series data is well-maintained in projected vector space.  Such results reflect that the semantics of time series can be accurately represented by their corresponding latent representations. We also observe that the temporal slice nearby in the vector space forms some successive sub-sequences (a.k.a. shapelets) in raw time series. This phenomenon indicates that the embedding learned by the FormerTime retains the potential of preserving the strength of shapelet-based methods. 
\subsubsection{Visualization of the Extracted Embeddings}
\begin{figure}
	\centering
	\includegraphics[width=0.45\textwidth]{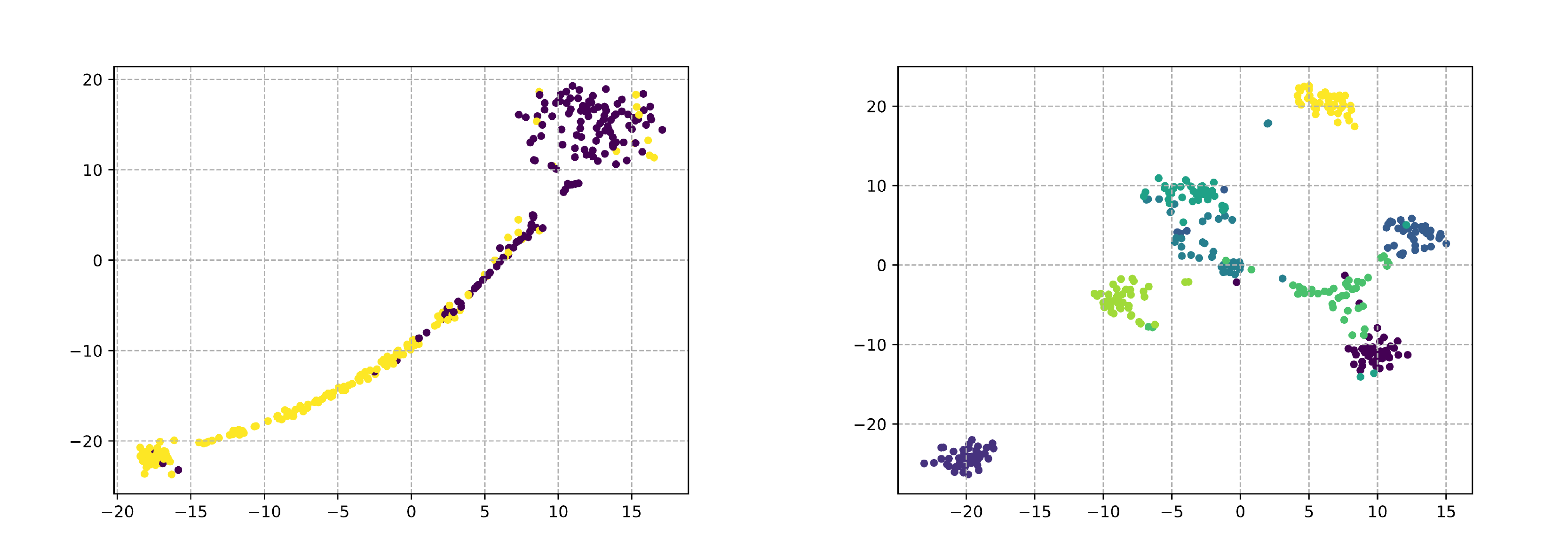} 
	\vspace{-0.1in}
	\caption{Visualization of the representation of whole time series on the SRS1 (left plot) and UW (right plot) datasets, extracted by pooling operation from the last hidden layer.}
	\vspace{-0.1in}
	\label{fig:tsne}
\end{figure}

As shown in Figure~\ref{fig:tsne}, we visualize the extracted feature vector from FormerTime by applying t-SNE to reduce the dimension. Here, we randomly choose examples from SRS1 and UWG datasets. In this figure, each point denotes an example and the same color denotes the corresponding original class labels. This figure suggests that the proposed FormerTime is able to project the data into an easily separable space to ensure good classification results.

\section{Conclusion}
In this work, instead of employing the prevalent convolutional architecture as the main backbones, we proposed FormerTime, a hierarchical transformer network for MTSC tasks. In FormerTime, both the strengths of transformers and CNNs were well absorbed in a unified model for further improving the classification capacity. Specifically, two aspects of vital inductive bias, including multi-scale time series representations and temporal-invariant capacity, are incorporated for enhancing the classification capacity in the MTSC task. Moreover, the terrible computation dilemma incurred by the self-attention mechanism was largely overcome in the proposed FormerTime, whose computation costs are acceptable for very long-sequence time series and large-scale data. Extensive experiments conducted on $10$ publicly available datasets from the UEA archive demonstrated that FormerTime could surpass previous strong baseline methods. In the future, we hope to empower the transferability of FormerTime~\cite{cheng2021learning}.

\begin{acks}
This research was partially supported by grants from the National Key Research and Development Program of China (No. 2021YFF0901003)
\end{acks}

\bibliographystyle{ACM-Reference-Format}
\bibliography{formertime_www23_camera_ready}
\appendix

\end{document}